\documentclass{article}

\usepackage{microtype}
\usepackage{graphicx}
\usepackage{subcaption}
\usepackage{booktabs} 

\usepackage{hyperref}

\usepackage[preprint]{icml2026}

\usepackage{amsmath}
\usepackage{amssymb}
\usepackage{mathtools}
\usepackage{amsthm}

\usepackage[capitalize,noabbrev]{cleveref}

\theoremstyle{plain}

\theoremstyle{definition}

\theoremstyle{remark}

\icmltitlerunning{A Unifying Lens on Reward Uncertainty in RLHF}

\begin{document}

\twocolumn[
  \icmltitle{A Unifying Lens on Reward Uncertainty in RLHF}

  \icmlsetsymbol{equal}{*}

  \begin{icmlauthorlist}
    \icmlauthor{Ely Hahami}{college,equal}
    \icmlauthor{Yoel Zimmermann}{seas,equal}
    \icmlauthor{Ray Zhou}{college}
    \icmlauthor{Jack Benarroch Jedlicki}{seas}
  \end{icmlauthorlist}

  \icmlaffiliation{college}{Harvard College, Harvard University, Cambridge, MA, USA}
    \icmlaffiliation{seas}{John A. Paulson School of Engineering and Applied Sciences, Harvard University, Cambridge, MA, USA}

  \icmlcorrespondingauthor{Yoel Zimmermann}{yzimmermann@g.harvard.edu}

  \icmlkeywords{RLHF, Reward Hacking, Distributional Reward Models, Pessimism, KL-DRO}

  \vskip 0.3in
]

\printAffiliationsAndNotice{\icmlEqualContribution}

\begin{abstract}
Reinforcement learning from human feedback (RLHF) is bottlenecked by \emph{reward hacking}, where the policy exploits errors in a proxy reward model (RM) and produces high RM scores without genuine quality gains. A natural mitigation is \emph{pessimism}: lowering rewards in regions where the RM is uncertain. However, standard scalar RMs provide no principled notion of uncertainty. We argue that the right object is a \emph{distributional} reward model $p(r\mid x,y)$. Under either a Bayesian inference or a KL-distributionally robust optimization (KL-DRO) lens, the KL-regularized RLHF objective admits a closed-form effective reward $\tilde r(x,y) = \pm\beta\log\mathbb{E}_p[e^{\pm r/\beta}]$. The pessimistic branch unifies the prior heuristics for RM ensemble aggregation: mean aggregation, worst-case optimization (WCO), and uncertainty-weighted optimization (UWO) all emerge as limits or truncations of this single expression. This also clarifies the implicit assumptions of each existing rule.
\end{abstract}

\section{Introduction}\label{sec:intro}

Reinforcement learning from human feedback (RLHF) is the dominant paradigm for aligning large language models with user intent ~\citep{christiano2017deep,ouyang2022instructgpt,bai2022constitutional,stiennon2020learning}. Pairwise preferences are first used to fit a reward model (RM) and a policy is then optimized against the RM with a KL penalty to a reference policy, via policy-gradient methods such as PPO~\citep{schulman2017proximalpolicyoptimizationalgorithms} or GRPO~\cite{Shao2024DeepSeekMath}. Because the RM is an imperfect proxy for human preferences, policy optimization can overoptimize into out-of-distribution regions, producing high RM scores without improving true response quality. This is an instance of Goodhart's law \citep{gao2023scaling,skalse2022defining,amodei2016concrete,pan2022effects} and has been documented across forms including sycophancy and \citep{sharma2023sycophancy,perez2023discovering} length bias \citep{singhal2023long}. 

A natural mitigation is \emph{pessimism}: when the RM is uncertain, a lower effective reward should be assigned. Prior work in RLHF instantiates this idea by training ensembles of RMs and aggregating their outputs in different ways. \citet{coste2024reward} introduce three rules: \emph{mean aggregation} (averaging members), \emph{worst-case optimization} (WCO, taking the minimum across members), and \emph{uncertainty-weighted optimization} (UWO, mean minus a tunable multiple of the empirical variance). \citet{eisenstein2023helping} make a similar case for ensembles, and follow-up work has explored adversarial \citep{chakraborty2024advpo} and robustness-driven \citep{Yan2024Reward} variants. These methods empirically help, but are typically presented as heuristics with little principled guidance on which to prefer and how their hyperparameters relate.

We argue here that to properly handle uncertainty, one should replace the scalar RM with a \emph{distributional} reward model $p(r \mid x, y)$. Once the RM is distributional, a single principle---either marginalization over reward uncertainty through a Bayesian inference lens \citep{korbak2022rl,levine2018reinforcement} or KL-distributionally robust optimization (KL-DRO) \citep{ben2009robust,hansen2008robustness}---yields a closed-form effective reward to substitute into the standard KL-RL objective. The aggregation rules used in practice emerge as truncations or limits of this expression. This perspective also connects RLHF pessimism to distributional reinforcement learning \citep{bellemare2017distributional,dabney2018distributional,dabney2018implicit,keramati2020being} and to offline-RL pessimism \citep{jin2021pessimism,xie2021bellman,rigter2022rambo}.

\paragraph{Contributions.}
\begin{itemize}
\itemsep0em
\item \textbf{A unifying lens.} From either a Bayesian inference or KL-DRO perspective, the RLHF objective with reward uncertainty admits a closed-form effective reward $\tilde r(x,y) = \pm\beta\log\mathbb{E}_p[e^{\pm r/\beta}]$ (Sec.~\ref{sec:theory}).
\item \textbf{Unification of prior heuristics.} Mean, WCO, and UWO emerge as the $\beta\to\infty$ limit, $\beta\to 0$ limit, and Gaussian truncation, respectively, of the pessimistic branch (Sec.~\ref{sec:unification}).
\item \textbf{Implementation guidance.} We discuss what each heuristic implicitly assumes about the reward distribution $p(r\mid x,y)$, and what is needed (calibration, posterior families) to instantiate the framework with a real RM (Sec.~\ref{sec:implementation}).
\end{itemize}

\section{Background and Related Work}\label{sec:related}

\paragraph{Reward overoptimization.} \citet{gao2023scaling} established that the proxy RM score and a stronger ``gold'' RM score (which is used to model the true objective we seek to optimize, i.e., how a human would score the response) diverge as PPO progresses, the canonical signature of reward hacking. \citet{skalse2022defining} formalize reward hacking, and \citet{amodei2016concrete} situate it among broader AI safety concerns. Empirically, hacked policies exhibit sycophancy \citep{sharma2023sycophancy}, length inflation \citep{singhal2023long}, and other surface artifacts that the proxy RM cannot distinguish from true quality \citep{wang2024automatic}.

\paragraph{Mitigating reward hacking with pessimism.} \citet{coste2024reward} introduce Mean, WCO, and UWO aggregation over RM ensembles and find that pessimistic variants reduce reward hacking. \citet{eisenstein2023helping} reach similar conclusions for best-of-$n$ and PPO, while noting that ensembles can share failure modes. Lightweight ensembles share a frozen backbone and diversify only a head~\citep{Dwaracherla2024Efficient}. Weight averaging the members instead of ensembling them~\citep{rame2024warm} trades the distributional signal for robustness and is complementary to our analysis. Pessimism can also be built into the reward model itself, by fine-tuning it to be pessimistic~\citep{xu2025learningpessimisticrewardmodel}, or imposed through pessimistic
preference-based policy objectives~\citep{gupta2025mitigatingpreferencehackingpolicy}. \citet{xu2025ask} use RM uncertainty to route hard cases to a stronger LLM judge.

\paragraph{Distributional reward models.} An alternative to ensembles is to place a posterior directly over a single head. Bayesian last-layer methods \citep{yang2026rewarduq,zhai2023uncertainty} yield a closed-form Gaussian posterior over the reward $\mathcal{N}(\mu(x,y), \sigma^2(x,y))$ at each prompt-response pair. \citet{dorka2024quantileregressiondistributionalreward} learns a full, potentially multimodal reward distribution by quantile regression and optimizes its lower quantiles for risk-aware RLHF, and \citet{sun2025probabilisticuncertainrewardmodel} generalizes the
Bradley-Terry likelihood to output a per-pair reward distribution whose
uncertainty penalizes the policy. Our framework specifies how to use any
such distributional RM in the standard KL-regularized objective.

\paragraph{Control-as-inference and robust RL.} The Bayesian view of KL-regularized RL \citep{korbak2022rl,levine2018reinforcement} interprets policies as posteriors under an ``optimality'' observation. Robust optimization \citep{ben2009robust,hansen2008robustness} and pessimism in offline RL \citep{jin2021pessimism,xie2021bellman,rigter2022rambo} provide the complementary, adversarial framing. Our derivation shows that, in the RLHF setting, these two seemingly disparate views produce \emph{the same functional form} up to a sign.

\section{Theory: A Unifying Lens}\label{sec:theory}

\subsection{Setup}\label{sec:setup}

Standard KL-regularized RLHF \citep{stiennon2020learning,ouyang2022instructgpt} maximizes
\begin{equation}\label{eq:rlhf}
\max_{\theta}\;\mathbb{E}_{x \sim \mathcal{D}}\!\Big[\mathbb{E}_{y \sim \pi_\theta(\cdot \mid x)} [r(x,y)] - \beta\, D_{\mathrm{KL}}\!\left(\pi_\theta\|\pi_{\mathrm{ref}}\right)\!\Big],
\end{equation}
with the well-known optimum $\pi^*(y\mid x) \propto \pi_{\mathrm{ref}}(y\mid x)\exp(r(x,y)/\beta)$.

In the standard pipeline, $r(x,y) = r_\phi(x,y)$ is a deterministic scalar from a learned RM. We instead treat the reward as a random variable $r \sim p(r \mid x, y)$, arising, for example, from a deep ensemble or a Bayesian head, and ask: to best mitigate reward hacking, \emph{what scalar effective reward $\tilde r(x,y)$ should replace $r_\phi(x,y)$ in Eq.~\ref{eq:rlhf}?}

\subsection{Optimistic Branch: Bayesian Marginalization}

Following the control-as-inference view \citep{korbak2022rl,levine2018reinforcement}, we introduce a binary optimality variable $O \in \{0,1\}$ with
\begin{equation} \label{eq:inference_as_control}
p(O{=}1\mid x,y,r) = e^{r/\beta},
\end{equation}
where $r$ is shifted\footnote{For posteriors with unbounded support (e.g., Gaussians), the likelihood reading of Eq.~\ref{eq:inference_as_control} is informal, as is standard in control-as-inference \citep{levine2018reinforcement}. Eq.~\ref{eq:opt} remains well defined whenever the MGF exists.} so that $\sup r = 0$. Marginalizing over reward uncertainty gives
\begin{equation}
p(O{=}1\mid x,y) = \int e^{r/\beta} p(r\mid x,y)\,dr = M_x(1/\beta),
\end{equation}
where $M_x(t) := \mathbb{E}_p[e^{tr}]$ is the moment generating function of $p(r\mid x,y)$. Bayes' rule with prior $\pi_{\mathrm{ref}}$ then yields $\pi^*_{\mathrm{opt}}(y\mid x) \propto \pi_{\mathrm{ref}}(y\mid x)\, M_x(1/\beta)$. Matching to the KL-RL form $\pi^* \propto \pi_{\mathrm{ref}}\exp(\tilde r/\beta)$ identifies
\begin{equation}\label{eq:opt}
\boxed{\;\tilde r_{\mathrm{opt}}(x,y) = \beta \log \mathbb{E}_{p(r\mid x,y)}\!\left[e^{r/\beta}\right].\;}
\end{equation}
By Jensen's inequality, since $\exp$ is convex,
\begin{equation}
\tilde r_{\mathrm{opt}}(x,y) \;\geq\; \mu(x,y),
\end{equation}
so uncertainty \emph{inflates} the effective reward. This is the Bayesian-optimal use of an uncertain RM under the control-as-inference likelihood in Eq.~\ref{eq:inference_as_control} in the absence of any robustness concern.

\subsection{Pessimistic Branch: KL-DRO}

To mitigate reward hacking we instead want a \emph{worst-case} effective reward. We let an adversary choose the reward distribution $Q$, paying a KL cost to deviate from the believed posterior $p$ \citep{ben2009robust,hansen2008robustness}:
\begin{equation}\label{eq:dro}
\tilde r_{\mathrm{rob}}(x,y) = \inf_Q \!\Big\{\mathbb{E}_Q[r] + \beta\, D_{\mathrm{KL}}(Q\,\|\,p(r\mid x,y))\Big\}.
\end{equation}
The parameter $\beta$ controls the adversary's reach: large $\beta$ pins $Q$ close to $p$, small $\beta$ permits arbitrary tilting. Solving the variational problem (see Appendix~\ref{app:theory}) yields the unique optimizer
\begin{equation} \label{eq:tilted}
Q^*(r) \;\propto\; p(r\mid x,y)\, e^{-r/\beta},
\end{equation}
which is $p$ exponentially tilted toward low rewards. Plugging back in,
\begin{equation}\label{eq:rob}
\boxed{\;\tilde r_{\mathrm{rob}}(x,y) = -\beta \log \mathbb{E}_{p(r\mid x,y)}\!\left[e^{-r/\beta}\right].\;}
\end{equation}
This is the \emph{entropic risk measure}~\citep{follmer2025stochastic} of $p(r\mid x,y)$. By Jensen's inequality, $\tilde r_{\mathrm{rob}}(x,y) \leq \mu(x,y)$: uncertainty \emph{deflates} the effective reward.

\paragraph{Choice of KL coefficients.} The KL coefficient $\beta$ in Eq.~\ref{eq:rlhf} controls the deviation of the policy from the reference, while $\beta$ in Eq.~\ref{eq:dro} controls the robustness to reward uncertainty. These play distinct roles and need not coincide. In this work we tie the two for simplicity, yielding the simple coefficient $\lambda=1 /(2 \beta)$. More generally one may consider $c \beta$, with $c$ reflecting the desired degree of pessimism relative to policy regularization.

\subsection{Cumulant Expansion}\label{sec:cumulants}

Let $K_x(t) := \log M_x(t) = \log\mathbb{E}_p[e^{tr}]$ be the cumulant generating function (CGF) of $p(r\mid x,y)$, with cumulants $\kappa_n$ defined by $K_x(t) = \sum_{n\geq 1} \kappa_n t^n / n!$. Then $\kappa_1 = \mu$, $\kappa_2 = \sigma^2$, $\kappa_3$ is the third central moment, $\kappa_4 = \mathbb{E}[(r-\mu)^4] - 3\sigma^4$ (excess kurtosis $\times\sigma^4$), and so on.

Substituting into Eqs.~\ref{eq:opt} and~\ref{eq:rob}:
\begin{equation}\label{eq:cumulants}
\tilde r_{\mathrm{opt/rob}} = \mu \pm \frac{\sigma^2}{2\beta} + \frac{\kappa_3}{6\beta^2} \pm \frac{\kappa_4}{24\beta^3} + \cdots
\end{equation}
Even cumulants ($\kappa_2,\kappa_4,\ldots$) flip sign between the two while odd cumulants ($\kappa_3,\kappa_5,\ldots$) do not. For symmetric distributions ($\kappa_{2n+1} = 0$ for $n\geq 1$), $\tilde r_{\mathrm{opt}}$ and $\tilde r_{\mathrm{rob}}$ are exact mirror images about $\mu$.

\paragraph{Gaussian case.} If $p(r\mid x,y) = \mathcal{N}(\mu, \sigma^2)$, all cumulants beyond $\kappa_2$ vanish, $K_x(t) = \mu t + \sigma^2 t^2 / 2$, and the series in Eq.~\ref{eq:cumulants} terminate exactly:
\begin{equation}\label{eq:gauss}
\tilde r_{\mathrm{opt}} = \mu + \frac{\sigma^2}{2\beta},\qquad \tilde r_{\mathrm{rob}} = \mu - \frac{\sigma^2}{2\beta}.
\end{equation}
Under any Gaussian distributional RM, the principled pessimistic reward is $\mu - \sigma^2/2\beta$, with the variance coefficient \emph{set by the KL coefficient $\beta$} rather than tuned as a free hyperparameter.

\begin{figure}[t]
\centering
\includegraphics[width=\columnwidth]{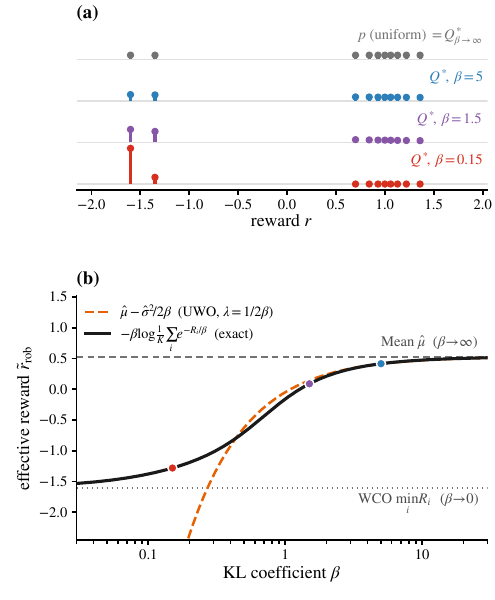}
\caption{\textbf{The pessimistic effective reward interpolates between Mean and WCO.}
Illustration for a $K{=}10$ RM ensemble that is bimodal on a suspect response
(eight members assign high reward, two flag it).
\textbf{(a)}~The adversarially tilted distribution $Q^*(r)\propto p(r)\,e^{-r/\beta}$
(Eq.~\ref{eq:tilted}): as $\beta$ decreases, the adversary concentrates mass on the worst members.
\textbf{(b)}~The exact effective reward $\tilde r_{\mathrm{rob}}$ (Eq.~\ref{eq:rob}) recovers
Mean as $\beta\to\infty$ and WCO as $\beta\to 0$ (Sec.~\ref{sec:unification}). The Gaussian truncation
$\hat\mu-\hat\sigma^2/2\beta$ (UWO with $\lambda{=}1/2\beta$) agrees at large $\beta$
but is over-pessimistic at small $\beta$, dropping below $\min_i R_i$ which no
distribution supported on the ensemble can justify (Sec.~\ref{sec:estimation}).}
\label{fig:unification}
\end{figure}

\subsection{Unification of Prior Methods}\label{sec:unification}

The aggregation rules of \citet{coste2024reward} emerge as limits and truncations of $\tilde r_{\mathrm{rob}}$.

\paragraph{Mean ($\beta \to \infty$).} As $\beta \to \infty$, the variance term in Eq.~\ref{eq:cumulants} vanishes ($\sigma^2/2\beta \to 0$), and similarly for all higher orders. Hence $\tilde r_{\mathrm{rob}} \to \mu$. Equivalently, an infinite KL penalty forces the adversary's distribution $Q^*$ back onto $p$, so the worst case coincides with the mean. \textbf{Mean aggregation is the risk-neutral limit of the pessimistic effective reward.}

\paragraph{WCO ($\beta \to 0$).} As $\beta \to 0$, the tilted distribution $Q^*(r) \propto p(r) e^{-r/\beta}$ concentrates all of its mass on $\inf\,\mathrm{supp}(p)$. For an empirical ensemble $\{R_i\}_{i=1}^K$ where $p$ is a uniform mixture of point masses, the infimum is $\min_i R_i$. Hence
\begin{equation}
\lim_{\beta \to 0} \tilde r_{\mathrm{rob}}(x,y) \;=\; \min_i R_i(x,y).
\end{equation}
\textbf{Worst-case optimization is the unbounded-adversary limit of the pessimistic effective reward.}

\paragraph{UWO (Gaussian truncation).} The UWO rule of \citet{coste2024reward} is
\begin{equation}
R_{\mathrm{UWO}}(x,y) = \hat\mu(x,y) - \lambda\,\hat\sigma^2(x,y),
\end{equation}
with $\lambda$ treated as a free hyperparameter. Intuitively, UWO works by penalizing the policy for generating responses for
which there is high disagreement among reward models within the ensemble, which helps prevent the
exploitation of a single reward model which might be erroneously assigning high rewards to
incorrect or low-quality responses. Comparing with the Gaussian case of Eq.~\ref{eq:gauss}, UWO is \emph{exactly} the pessimistic effective reward under a Gaussian posterior, with $\lambda = 1/(2\beta)$. UWO's implicit assumptions are: (i) $p(r\mid x,y)$ is approximately Gaussian and (ii) the variance coefficient can be decoupled from the KL coefficient $\beta$.

\begin{table}[t]
\caption{Existing pessimistic aggregation rules \citep{coste2024reward} as limiting cases or truncations of $\tilde r_{\mathrm{rob}}$. The fourth row gives the principled default set by the KL coefficient $\beta$ under a Gaussian posterior, with no extra hyperparameter.}
\label{tab:unification}
\vskip 0.05in
\begin{center}
\begin{small}
\begin{sc}
\begin{tabular}{@{}p{1.5cm}@{\hskip 4pt}p{2.2cm}@{\hskip 4pt}p{3cm}@{}}
\toprule
Method & Formula & Recovered as \\
\midrule
Mean & $\hat{\mu}$ & $\beta\to\infty$ limit \\
WCO  & $\min_i R_i$ & $\beta\to 0$ limit \\
UWO  & $\hat{\mu}-\lambda\hat{\sigma}^2$ & Gaussian, $\lambda$ free \\
\textbf{Ours} & $\hat{\mu}-\hat{\sigma}^2/2\beta$ & Gaussian, $\lambda=1/2\beta$ \\
\bottomrule
\end{tabular}
\end{sc}
\end{small}
\end{center}
\vskip -0.15in
\end{table}

Table~\ref{tab:unification} and Figure~\ref{fig:unification} summarize the unification. The take-away is that the choice of aggregation rule is a choice about the assumed shape of $p(r\mid x,y)$ and the adversary's KL budget. Mean assumes no uncertainty matters, WCO assumes the worst possible distribution, and UWO is the middle ground in a Gaussian approximation. Our framework makes these assumptions explicit and supplies a principled default ($\lambda = 1/2\beta$) when working with a Gaussian distributional RM.

\subsection{Estimation from a Finite Sample}\label{sec:estimation}

In practice $p(r\mid x,y)$ is approximated by a finite sample $\{r_i(x,y)\}_{i=1}^K$, either ensemble members or posterior samples:
\begin{equation}
\hat\mu = \frac{1}{K}\sum_{i=1}^K r_i,\quad \hat\sigma^2 = \frac{1}{K-1}\sum_{i=1}^K (r_i - \hat\mu)^2.
\end{equation}
Two natural estimators of $\tilde r_{\mathrm{rob}}$ arise.

\paragraph{Direct log-MGF estimator.} 
\begin{equation}
\hat{\tilde r}_{\mathrm{rob}}^{\,(\text{logMGF})} = -\beta\log\!\left(\frac{1}{K}\sum_{i=1}^K e^{-r_i/\beta}\right).
\end{equation}
This estimator is consistent but biased at finite $K$: since $\log$ is concave, Jensen gives $\mathbb{E}[\hat{\tilde r}_{\mathrm{rob}}^{\,(\text{logMGF})}] \geq \tilde r_{\mathrm{rob}}$, with bias growing in $\sigma/\beta$. It is sensitive to outliers when $\beta$ is small (the $\exp(-r_i/\beta)$ term blows up for any single low $r_i$).

\paragraph{Gaussian-truncated estimator.}
\begin{equation}\label{eq:est-gauss}
\hat{\tilde r}_{\mathrm{rob}}^{\,(\text{Gauss})} = \hat\mu - \frac{\hat\sigma^2}{2\beta}.
\end{equation}
This is exact when $p$ is Gaussian and drops higher cumulants otherwise. We note that for small $K$ (as is typical for RM ensembles, $K \in \{3,\ldots,10\}$ in \citet{coste2024reward}), higher cumulants are poorly estimated. The relative error on $\hat\sigma^2$ alone is $\sqrt{2/(K-1)} \approx 71\%$ for $K=5$, so we recommend Eq.~\ref{eq:est-gauss} as a default.

\section{Discussion}\label{sec:implementation}

The pessimistic effective reward $\tilde r_{\mathrm{rob}}$ is only as informative as the posterior $p(r\mid x,y)$ used to compute it. Two concrete prerequisites:

\textbf{(i) A genuine distributional RM.} Scalar RMs trained with the Bradley-Terry likelihood \citep{bradley1952rank} provide no uncertainty. Two practical options are: (a) a deep ensemble \citep{Lakshminarayanan2016Simple,coste2024reward,Dwaracherla2024Efficient}, whose empirical variance approximates $\sigma^2(x,y)$; and (b) a Bayesian last-layer construction \citep{yang2026rewarduq,zhai2023uncertainty}, which yields a closed-form Gaussian posterior at the cost of a single forward pass and an $O(d^2)$ matrix-vector product. The Bayesian construction is especially interesting as its posterior is Gaussian and Eq.~\ref{eq:gauss} applies without truncation.

\textbf{(ii) Calibration.} The variance term $\sigma^2(x,y)$ only modulates $\tilde r_{\mathrm{rob}}$ usefully if it tracks true epistemic uncertainty. The Bradley--Terry loss collapses soft preferences ($0.51$ and $0.99$) onto the same hard label and degrades calibration. Strictly proper scoring rules \citep{gneiting2007strictly}, in particular the Brier score \citep{brier1950verification}, preserve annotator confidence and are known to improve calibration. Using AI feedback \citep{bai2022constitutional,lee2024rlaifvsrlhfscaling} provides soft labels naturally by prompting for a confidence score during AI feedback. 

\paragraph{Implications for existing methods.}
\begin{itemize}
\itemsep0em
\item \emph{When Mean should suffice.} The cumulant expansion (Eq.~\ref{eq:cumulants}) shows that pessimism corrections decay as $1/\beta^{n-1}$. For large $\beta$ (heavy KL regularization), Mean is essentially optimal as all rules collapse to it.
\item \emph{When WCO is appropriate.} WCO is the $\beta \to 0$ limit, suitable when the adversary is essentially unconstrained. In RLHF, this corresponds to a setting where one expects the RM to be arbitrarily wrong in some direction. \citet{eisenstein2023helping} report that WCO can underperform Mean when the ensemble is too small or correlated, which our framework explains: WCO discards all distributional information beyond the minimum.
\item \emph{Why UWO works.} UWO is the right shape under a Gaussian posterior. Its empirical success suggests that practical RM ensembles are reasonably symmetric and unimodal. The free $\lambda$ may be absorbing miscalibration; under the theory-prescribed value $\lambda = 1/2\beta$, calibration becomes essential.
\end{itemize}

\paragraph{Beyond Gaussian.} The cumulant expansion in Eq.~\ref{eq:cumulants} suggests three directions when $p(r\mid x,y)$ is not well-approximated by a Gaussian. (i) For asymmetric but light-tailed distributions, third- and fourth-cumulant corrections to Eq.~\ref{eq:est-gauss} can be estimated from larger ensembles. (ii) For genuinely heavy-tailed distributions where the MGF estimator is unstable, quantile-based pessimism (e.g., CVaR) is a robust alternative, connecting to quantile distributional RL \citep{dabney2018distributional,dabney2018implicit,keramati2020being}. (iii) Flexible posterior families, such as mixtures of Gaussians or normalizing flows fit directly to held-out preference data could replace the ensemble or Bayesian linear head.

\paragraph{Connections to broader pessimism.} The KL-DRO derivation connects RLHF reward hacking to pessimism in offline RL \citep{jin2021pessimism,xie2021bellman,rigter2022rambo}, where similar log-MGF / entropic-risk formulations arise from analogous principles. Our contribution is to localize this connection: in RLHF, the relevant random variable is the \emph{reward} (not the transition dynamics, which are deterministic for an autoregressive policy), and the relevant penalty parameter is the standard KL coefficient $\beta$ already present in Eq.~\ref{eq:rlhf}.

\section{Conclusion}\label{sec:conclusion}

We have given a single derivation that captures the essence of reward uncertainty in RLHF. From either a Bayesian or a KL-DRO starting point, the closed-form effective reward is $\tilde r(x,y) = \pm\beta\log\mathbb{E}_p[e^{\pm r/\beta}]$. The pessimistic branch unifies mean aggregation, WCO, and UWO as limits and truncations, and yields a principled default $\tilde r = \mu - \sigma^2/2\beta$ under a Gaussian posterior in which the variance coefficient is set by the KL coefficient rather than tuned. The framework clarifies what each existing heuristic implicitly assumes and gives a recipe for going beyond Gaussian via higher cumulants or quantile-based pessimism.

\section*{Acknowledgements}
We thank Kiante Brantley for initial guidance and helpful discussions during the early stages of this project. OpenAI GPT-5.5 and Refine.ink were used to assist with language editing and improve the clarity and readability of the manuscript.
\bibliography{references}
\bibliographystyle{icml2026}

\newpage
\appendix
\onecolumn

\section{Proofs and Derivations}\label{app:theory}

\subsection{Optimistic Effective Reward}

\paragraph{Setup.} Let $\pi_0 = \pi_{\mathrm{ref}}$. For each $(x,y)$, $r \sim p(r\mid x,y)$. The KL-regularized objective with effective reward $\tilde r$ is
\begin{equation}
J(\pi) = \mathbb{E}_\pi[\tilde r(x,y)] - \beta\, D_{\mathrm{KL}}(\pi \| \pi_0), \qquad \beta > 0.
\end{equation}

\paragraph{Step 1: Optimality variable.} Introduce $O \in \{0,1\}$ with $p(O=1\mid x,y,r) = e^{r/\beta}$ where $r$ is shifted so $\sup r = 0$.

\paragraph{Step 2: Marginalize.} 
\begin{equation}
p(O{=}1\mid x,y) = \int e^{r/\beta} p(r\mid x,y)\,dr = M_x(1/\beta),
\end{equation}
where $M_x(t) := \mathbb{E}_p[e^{tr}]$ is the MGF.

\paragraph{Step 3: Bayes' rule.} With prior $\pi_0$,
\begin{equation}
\pi^*_{\mathrm{opt}}(y\mid x) \propto \pi_0(y\mid x) M_x(1/\beta).
\end{equation}

\paragraph{Step 4: Identify $\tilde r$.} Matching to $\pi^* \propto \pi_0 \exp(\tilde r/\beta)$ gives
\begin{equation}
\tilde r_{\mathrm{opt}}(x,y) = \beta \log \mathbb{E}_p[e^{r/\beta}].
\end{equation}

\paragraph{Step 5: Jensen.} Since $\exp$ is convex, $\mathbb{E}_p[e^{r/\beta}] \geq e^{\mu/\beta}$, so $\tilde r_{\mathrm{opt}} \geq \mu$.

\subsection{Pessimistic Effective Reward (KL-DRO)}

\paragraph{Step 1: Definition.}
\begin{equation}
\tilde r_{\mathrm{rob}}(x,y) = \inf_Q \{\mathbb{E}_Q[r] + \beta D_{\mathrm{KL}}(Q\|p)\}.
\end{equation}

\paragraph{Step 2: Lagrangian.} With multiplier $\lambda$ for $\int Q = 1$,
\begin{equation}
\mathcal{L}(Q,\lambda) = \int Q(r) r\,dr + \beta\!\int Q\log Q\,dr - \beta\!\int Q\log p\,dr - \lambda\!\left(\int Q\,dr - 1\right).
\end{equation}

\paragraph{Step 3: Stationarity.} Using $\frac{d}{dQ}[Q\log Q] = \log Q + 1$:
\begin{equation}
\frac{\delta \mathcal{L}}{\delta Q(r)} = r + \beta(\log Q + 1) - \beta\log p - \lambda = 0,
\end{equation}
giving $Q^*(r) \propto p(r) e^{-r/\beta}$. The normalization is $C(x,y) := \mathbb{E}_p[e^{-r/\beta}]$.

\paragraph{Step 4: Evaluate the infimum.} $\log Q^*(r) - \log p(r) = -r/\beta - \log C$, so
\begin{equation}
D_{\mathrm{KL}}(Q^*\|p) = -\tfrac{1}{\beta}\mathbb{E}_{Q^*}[r] - \log C.
\end{equation}
Therefore $\mathbb{E}_{Q^*}[r] + \beta D_{\mathrm{KL}}(Q^*\|p) = -\beta\log C$ (the reward terms cancel exactly), giving
\begin{equation}
\tilde r_{\mathrm{rob}}(x,y) = -\beta\log\mathbb{E}_p[e^{-r/\beta}].
\end{equation}

\paragraph{Step 5: Jensen.} $\mathbb{E}_p[e^{-r/\beta}] \geq e^{-\mu/\beta}$; taking $-\beta\log$ reverses the inequality, so $\tilde r_{\mathrm{rob}} \leq \mu$.\hfill$\square$

\subsection{Cumulant Expansion}

Let $K_x(t) = \log\mathbb{E}_p[e^{tr}] = \sum_{n\geq 1} \kappa_n t^n / n!$. Then
\begin{equation}
\tilde r_{\mathrm{opt}} = \beta K_x(1/\beta) = \sum_{n\geq 1} \frac{\kappa_n}{n!\,\beta^{n-1}} = \mu + \frac{\sigma^2}{2\beta} + \frac{\kappa_3}{6\beta^2} + \frac{\kappa_4}{24\beta^3} + \cdots,
\end{equation}
\begin{equation}
\tilde r_{\mathrm{rob}} = -\beta K_x(-1/\beta) = \sum_{n\geq 1} \frac{(-1)^{n+1}\kappa_n}{n!\,\beta^{n-1}} = \mu - \frac{\sigma^2}{2\beta} + \frac{\kappa_3}{6\beta^2} - \frac{\kappa_4}{24\beta^3} - \cdots.
\end{equation}
For Gaussian $p$, $\kappa_{n\geq 3} = 0$ and the series terminate at $\tilde r = \mu \pm \sigma^2/2\beta$.

\subsection{Recovery of Mean and WCO}

\paragraph{$\beta\to\infty$ (Mean).} Each term $\kappa_n/(n!\beta^{n-1})$ for $n\geq 2$ vanishes, leaving $\tilde r_{\mathrm{rob}} \to \kappa_1 = \mu$. Equivalently, $Q^*(r) \propto p(r) e^{-r/\beta} \to p(r)$ as $\beta\to\infty$.

\paragraph{$\beta\to 0$ (WCO).} For an empirical ensemble $\{R_i\}_{i=1}^K$ with $p(r\mid x,y) = \frac{1}{K}\sum_i \delta(r - R_i)$,
\begin{equation}
\tilde r_{\mathrm{rob}} = -\beta\log\!\left(\frac{1}{K}\sum_{i=1}^K e^{-R_i/\beta}\right).
\end{equation}
Let $R_{\min} = \min_i R_i$ and factor out $e^{-R_{\min}/\beta}$:
\begin{equation}
\tilde r_{\mathrm{rob}} = R_{\min} - \beta\log\!\left(\frac{1}{K}\sum_{i=1}^K e^{-(R_i - R_{\min})/\beta}\right).
\end{equation}

As $\beta \rightarrow 0$, all terms in the sum with $R_i>R_{\min }$ are exponentially suppressed; the sum approaches the count of minimizing members $m$ divided by $K$. The logarithm of this sum approaches the constant $\log (m / K)$, so the overall correction term scales as $O(\beta) \rightarrow 0$. Hence $\tilde r_{\mathrm{rob}} \to R_{\min} = \min_i R_i$, recovering WCO.

\paragraph{Gaussian truncation (UWO).} For $p = \mathcal{N}(\mu, \sigma^2)$, $K_x(t) = \mu t + \sigma^2 t^2/2$ exactly, so $\tilde r_{\mathrm{rob}} = \mu - \sigma^2/2\beta$. This matches $R_{\mathrm{UWO}} = \hat\mu - \lambda\hat\sigma^2$ with $\lambda = 1/(2\beta)$.

\end{document}